\documentclass{IEEEtran} 
   
\usepackage{url}
\usepackage[hidelinks]{hyperref}
\usepackage[utf8]{inputenc}
\usepackage{graphicx}
\usepackage{amsmath}
\usepackage{booktabs}
\usepackage{algorithm}
\usepackage{algpseudocode}

\usepackage{color}
\usepackage{multirow}
\usepackage{xr}
\usepackage{listings}

\title{On the Power of Refined Skat Selection}

\author{
Stefan Edelkamp, AI Center at CTU Prague, Czech Republic
}

\begin{document}

\maketitle

\begin{abstract}
Skat is a fascinating combinatorial card game,
show-casing many of the intrinsic challenges for
modern AI systems such as cooperative and adversarial
behaviors (among the players), randomness (in
the deal), and partial knowledge (due to hidden
cards). Given the larger number of tricks and higher 
degree of uncertainty, reinforcement learning 
is less effective compared to classical board games
like Chess and Go.
As within the game of Bridge, in Skat we
have a bidding and trick-taking stage. 
Prior to the trick-taking and as part of
the bidding process, one phase in the game is to
select two skat cards, whose quality may
influence subsequent
playing performance drastically. 
This paper looks into different skat selection strategies. 
Besides predicting the probability of winning and 
other hand strength
functions we propose hard
expert-rules and a scoring functions based on refined
skat evaluation features. Experiments
emphasize the impact of the refined skat putting 
algorithm on the playing performance of the bots, 
especially for AI bidding and AI game selection.    
\end{abstract}

\lstset{language=C++,morekeywords={array,constraint,var,forall,sum,solve,minmize,decreasing,domain},basicstyle=\ttfamily,keywordstyle=\color{blue}\ttfamily,
literate=%
  {+}{{{\color{red}+}}}1
  {!}{{{\color{red}!}}}1
  {*}{{{\color{red}*}}}1
  {/}{{{\color{red}/}}}1
  {=}{{{\color{red}=}}}1
  {|}{{{\color{red}|}}}1
  {\%}{{{\color{red}$\%$}}}1
  {<}{{{\color{red}<}}}1
  {~}{{{\color{red}$\sim$}}}1
  {\&}{{{\color{red}\&}}}1 
  }

\newtheorem{definition}{Definition}
\newtheorem{theorem}{Theorem}
  \setlength{\tabcolsep}{5pt} 

\section{Introduction}

Many board games like Checkers~\cite{checkers}, Nine-Men-Morris~\cite{ninemenmorris} have been solved already, or, as in Chess or Go, computer AIs are clearly outperforming humans~\cite{AlphaGo,alphazero}.
After some variants of Poker have also been solved to a satisfying degree~\cite{poker},
Skat~\cite{Wergin} as well as Bridge~\cite{GIB} have been identified as 
two of the current game playing challenges in modern AIs~\cite{DBLP:journals/corr/abs-1911-07960,DBLP:conf/ijcai/BuroLFS09}, given that expert human card players still play consistently better than the machines. 

We consider Skat~\cite{Rainer}, as with three players and 32 cards it is more concise than Bridge, and still full of virtue. From a combinatorial search perspective, Skat has
$n={32 \choose 10}{22\choose 10}{12 \choose 10} \approx 2.8$ quadrillion possible deals. Following the well-known
\emph{birthday paradoxon}, the probability $p$ that
two deals appear twice is 
$p = 1 - (\prod_{i=0}^{k-1} (n-i)/(n^k))$. 
Using $p \ge 50\%$ this yields 
$k \ge 40$ million games. This easily cover the lifetime of a player. 

Skat is a prototype to illustrate the intrinsic difficulties in handling randomness, incomplete information, partial observability, epistemic reasoning as well as cooperation in adversarial and competitive multi-agent search scenarios.

As part of the bidding stage, right after taking the skat cards from the table, and right before announcing the game to be played and trick-taking, the declarer selects and discards two of twelve cards from his/her hand to be put. As the name of the game suggests, choosing these two skat cards is critical for all stages of the game, and deserves the special attention given to in this paper. 
We will see that selecting (taking and putting) skat is crucial to influence the overall playing performance of the game. With an unfortunate choice skat cards a game is often doomed to be lost, while it may be won with a slightly better one. 
For a ﬁxed game, there are ${12 \choose 2} = 66$ options to discard any $2$ of the $12$ cards. For selecting the game and each one to be played, one samples all ${22 \choose 2} = 231$ hands the skat cards to be taken may have, and computes the mean. For the bidding stage $b$ (required and confirmed), therefore, at each stage of bidding process 
$t$ games are selected and remaining hand strength such as winning probabilities are compared, resulting in considering 
${22 \choose 2} {12 \choose 2} tb$
skats. Before a game is declared, for selecting the best skat cards, refined reasoning is required. 
Therefore, the core contribution of this paper is the proposal of a refined \emph{skat selection strategy}, which suggests the two cards to be put and influences all stages of the games. 

The paper is structured as follows. After introducing the game of Skat
and looking into related work, we start building better and better skat selection options. We integrate the refined bidding system into a known player. Server and replaying experiments show that our efforts for improved skat selection pay off to increase the performance of the bots.

\section{About Skat}

Skat is a three-player imperfect information game played with 32 cards, a subset of the usual 52 cards Bridge deck.
It shares similarities to Marias(ch) (played in Czech Republic and Slovakia)
and Ulti (played in Hungary).

\begin{figure}
    \centering
    \includegraphics[width=8.5cm]{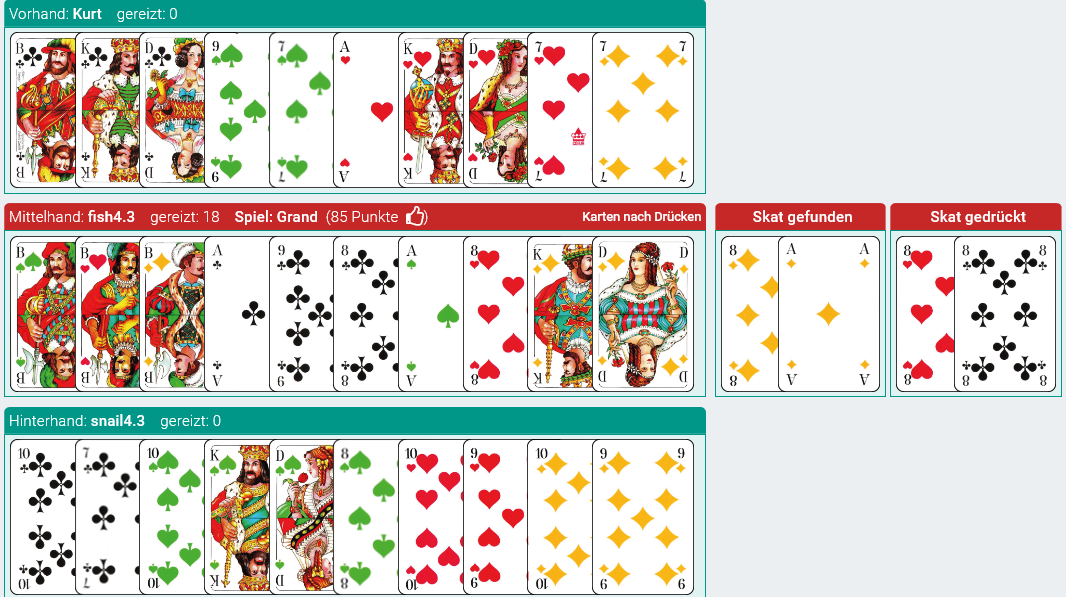}
    \caption{Skat deal in a game of a human against two AIs, with $\diamondsuit 8$ and $\diamondsuit$ A in the skat, and the eights in $\heartsuit$ and $\clubsuit$ being selected by the declarer bot
    for being discarded. }
    \label{fig:skatput}
\end{figure}

At the beginning of a game, each player gets 10 cards, which are hidden to the other players. The remaining two cards, called the skat, are placed face down on the table. Each hand is played in two stages, bidding and card play. 

The bidding stage determines the declarer and the two opponents: two players announce and accept increasing bids until one passes. 
The winner of the ﬁrst bidding phase continues bidding with the third player. 
The successful bidder of the second bidding phase plays against the other two. The maximum bid a player can announce depends on the type of game the player wants to play and, in case of a trump game, a multiplication factor determined by the jacks. The solist decides on the game to be played. Before declaring, he may pick up the skat and then discard any two cards from his hand, face down. 
These cards count towards the declarer's score. 
An example for skat selection is shown in Fig~\ref{fig:skatput}.

Card play proceeds as in Bridge, except that the trumps and card ranks are different. In grand,the four jacks are the only trumps. In suit, additional seven other cards of the selected suit are trumps. There are no trumps in null games. Non-trump cards are grouped into suits as in Bridge. Each card has an associated point value between 0 and 11, and in a standard trump game the declarer must score more points than the opponents to win. Null games are an exception, where the declarer wins only if he scores no trick.

Most games simply cannot be won without picking up the skat. 
A standard trump game is won with 61 points, (or eyes, as some skat players like to say). There are higher contracts with 90 (Schneider) and 120 (Schwarz)
games, which we consider but have dropped from the 
setting of this paper, mainly for the sake of simplicity. 
In hand games the skat is
not included to the hand for being put, but directly and uncovered
goes to the the own stack. This prevents us from computing the 
best skat evaluation score and take the mean. 

In any case, during bidding and game selection,
instead of distilling the best skat, a combination of the average and maximum evaluation over the skats is taken.

Considering the higher card values that count in favor to the declarer, it can be decisive for the outcome of the game, whether s/he puts 20 eyes, or 0 eyes into the skat. Putting trump ace and 10 to into the skat, just to have many points as possible in the skat, however, is strongly not recommended.

\section{Skat Selection Strategies}

Skat selection is much about distilling 
an estimate for the strength of a 
hand, predicting the possible outcome of the game and the subsequent payoff.

\subsection{Influence of Skat Selection}

For the sake of brevity, we assume no hand or ouvert game. We notice that improving the selecting of the two discard cards has immediate consequences for the other stages of the game, bidding as skat putting influences the bidding, trick-taking, as skat putting influences the knowledge the opponents will infer about the declarer's hand. In the following, we will derive an evaluation function $f(h,s)$ for the skat cards $s$ for a given hand $h$, and to select the skat that has the best score $\mbox{max}_{c,c' \in h \cup s, c \neq c'} f(h \cup s \setminus \{c,c'\})$. For each type of game
$t \in \{grand,null,\clubsuit,\spadesuit,\heartsuit,\diamondsuit\}$, we iterate 
over all possible ${12 \choose 2}= 66$ skats.

To determine the declarer during bidding we iterate over all possible skats \emph{for every bidding vector} $(b_0,b_1,b_2)$ with $b_i \in \{ 18,20,22,23,24 \ldots\}, i \in \{0,1,2\}$, as the current maximum bid $b = \max\{b_1,b_2,b_3\}$ has an influence on the probability of winning: if the other players
aim high in their bits, this usually indicates opponent strength, so that the bidder may rethink and lower his expectation. 

There are some general rules for good skat putting~\cite{Rainer},
such as (we abbreviate A for Ace, J for Jack, Q for Queen, K for King, and N for 7, 8, or 9):

\begin{itemize}
    \item put all eyes that otherwise would be 
    likely lost;
    \item in suit play with six trumps or more try to increase the number of free suits to be able to cut;
    \item in grands with few Js, keep as 
    many retaking opportunities as possible 
    even in case of a once-supported 10;
    \item keep suits with standing cards to win more tricks;
    \item take care of the position of the players: e.g., in rear hand, prefer free suits to once-supported 10s, while in fore- and middle-hand, the opposite rule 
    counts;
    \item In the combination A, 10, K (and maybe others) put the 10 to secure many eyes.
\end{itemize}

For standing cards, we distinguish emph{100\% standing cards}: cards in a suit, which in a minigame of that
tricks in a minigame to be won, given that 
the declarer has a sufficient number of Js or As to come back if leaving the trick; $iii)$ \emph{standing cards without or with retaking}: expected number of 
tricks in that minigame that will be won with or without having a sufficient number of leading cards to be able to take back 
the issuing right when leaving the trick.

As an example in standard grands with at least two Js, we have the following set of simple skat selection rules: 

\begin{itemize}
    \item try keeping the 100\% standing cards in hand;
    \item if 2~tricks have to be given away, a game can only be lost if the declarer provides two high cards, so for this case we prefer putting 10s;
    \item when losing 3~tricks, having  2~high cards at an own disposal, the game can only be lost, if Qs and Ks are provided, so that these should be discarded from the hand;
    \item if there are 4~tricks to be given away, and four high cards in the own hand or in the skat, drop all Qs and Ks;
    \item for 5~tricks to loose, the hand should be handled as a high-card game. 
\end{itemize} 
 
In the following we describe existing 
selection strategies.

\subsection{Random}

The simplest and clearly worst selection strategy is to choose two random cards to be discarded from the hand. This is only used as a baseline for comparison.

\subsection{Von Stegen}

There is a proposal to evaluate the strength of the given hand using the points system of Walter von Stegen, an expert Skat player, youtuber and teacher. So far, it only applies to trump games. Roughly speaking, it counts the number of Js with 2 points, and the number of trump and high-cards (As or 10s) with 1 point (high trump cards are counted twice). Additionally,
strong Js and low bids are awarded. We
choose the following implementation with bitvectors for card sets.

\begin{scriptsize}
\begin{lstlisting} 
vonStegen
 v = |hand & JACK|
 v += |hand & (ACE|10)|
 if (trumpmask != JACK)   
   v += |hand & trumpmask|
   if (|hand & CJ| = 0) 
    v += 0.5 * (|JACK & hand| > 2)
  v += 0.5 (|(CJ | SJ) & hand| = 2)
  v += 1.0 (|(CJ | SJ | HJ) & hand| = 3)
  v += 0.5 (|JACK & hand| = 4) 
  v += 0.5 (bidval = 0)
 return v;
\end{lstlisting}
\end{scriptsize}

\subsection{Kinback}

Thomas Kinback is a professional Skat player, teacher,
and author~\cite{Kinback}. He developed another counting system for hand strength, which is explained in greater depth by~\cite{Rainer}.
The hand strength value differs in grand and suit games, 
and includes a measure for tricks going home. Moreover, different 
configurations of cards in a suit as well as being in forehand get a surplus. We derived the following implementation for 
the system.

\begin{scriptsize}
\begin{lstlisting} 
kinback  
 if (trumpmask = JACK)  
  return (pos = 1) + |hand & ACE| + |hand & JACK|
 kb = 0;
 st = safetricks(hand & trumpmask);
 for (suit=0; suit < 4; suit++) 
  if (suit != trumpsuit) 
   h = hand & mask[suit]
   AT = 2.0 |h & (ACE|10)| = 2
   AK = 1.5 |h & ~10 & (ACE|KING)| = 2
   SA = 1.0 |h & ~10 & ~KING & ACE| = 1
   TK = 1.0 |h & ~ACE & (10|KINGS)| = 2
   T = .5 |h & ~ACE & (10|QUEEN|N)) = 3
   K = .5 |h & ~(ACE|10) & (KING|QUEEN|N)|=3
   kb += AT + AK + SA + TK + T3 + K3
  return .5 (pos = 1) + st + kb
\end{lstlisting}
\end{scriptsize}

\subsection{Winning Probability}

Edelkamp~\cite{DBLP:conf/ecai/Edelkamp20} provides a table estimation of the winning probability for a hand based on suits (null), or on on winning parameters (trump). For null games, the statistically sampled winning ratio $W(h,c)$ for each suit $c$ is used to estimate winning probability $W(h) = W(h,\clubsuit) \cdot W(h,\spadesuit) \cdot W(h,\heartsuit) \cdot W(h,\diamondsuit)$ of the hand $h$~\cite{Lasker,Lasker2}.
For trump games, we consider a hash table addressed by the so-called \emph{winning parameters}, such as 
\begin{itemize}
    \item $w_1$: number of non-trump suits that the player lacks;
\item $w_2$: number of eyes put in skat, condensed into 4 groups;
\item $w_3$: value of the bidding stage, projected to 4 groups;
\item $w_4$: position of the declarer in the first trick;
\item $w_5$: number of trump cards in hand;
\item $w_6$: number of non-trump cards in hand;
\item $w_7$: jacks groups; and 
\item $w_8$: number of trick estimated to loose; 
\end{itemize}
The estimates are derived from the results of millions of expert games and stored in two times two lookup hash tables, one fore- and one background table for suit, and the same two for grand games. The small background tables apply to a smaller set of winning parameters and are addressed in case the number of samples is not sufficient to derive a robust score in the large foreground one. The combination turns out to be rather accurate (on average at most 1\% off from the correct scoring result).

If the estimates of winning probabilities were exact (and fast enough to be computed/retrieved), skat selection would be a rather simple process. 
The problem, however, is that some winning parameter settings fail to have sufficiently many games
for a statistically relevant result, so that the robust and confident partitioning of all games along winning parameters is still not completely solved.
Grand games contribute about 28-30\% of the games. Null games 
about 7-8\%, which led to the different implementations.
 
As the various kinds of games have different game values during the bidding and accounting, usually evaluated in the Seeger scoring system, 
one translate winning probabilities $W^t(h)$ into
expected cost $C^t(h)$ for a given hand $h$ and game type $t$ using the game value $V^t(h)$ as follows 
\begin{small}
\begin{equation}\label{eq:eval}
C^t(h) = (50 +V^t(h)) W^t(h) 
 - (50 +2V^t(h)) (1-W^t(h)).
\end{equation}
\end{small}
For each of the 66 skat cards $s$ taken 
to the hand $h$ in game $t$ the winning probability would be determined, and the game $t \in\{grand,null,\clubsuit,\spadesuit,\heartsuit,\diamondsuit\}$ and according skat 
with highest expected return  
$$\mbox{argmax}_{c,c' \in h \cup s. c\neq c'} C^t(h \cup s \setminus \{c,c'\})$$
be chosen. While having derived an very accurate estimate for the
chance of winning before trick-taking play is a big achievement, it turns out to
be still too coarse to rank the different skats. 
Non-surprisingly, 
reducing skat selection 
to reading out a big probability 
table is an over-simplification.

\section{Skat Selection Refinements}

In the experiments we have noticed that neither of the above systems is sufficiently strong to determine a sufficiently good 
scoring function for ranking the skats 
that is able to beat expert human 
play on their skat selection. As \cite{DBLP:conf/socs/Edelkamp19}
found human-superior skat putting for null games,
our refinements tackle trump games.

\subsection{Hard Constraints Rules}

There are many possible skats inferior to others that should be neglected. For example, 
in a trump game it is, 
up to very rare exceptions, unwise to 
discard a trump card. One can easily come up with
a series of  rules to reduce the set
of plausible skats down from 66 significantly to about  
5-20 skats to be subsequently scored and ranked.

For example, in a high-card grand, we permit Js to be discarded, and 
enforce the putting of sole 10s, unless there
are three of them. If an A
and a 10 is put and there is a no K in the
same suit of length 3, this option is prevented from putting. If a 10 and A are available in one suits, and one is put, we prefer putting the A (one additional eye).

There is one further exception in grand, which
enforces skat cards following the 
\emph{high-card theorem}~\cite{Rainer,DBLP:conf/socs/Edelkamp19}:
\emph{if the number of high-valued cards secured by the declarer is at
least as large as the number of tricks lost (assuming no points
were put into them), he wins}. The implementation
not only checks for the theorem, but in case it holds, also
provides the skat that fulfils its preconditions.

In a standard grand, we permit discarding Js 
and As, and enforce putting of sole 10s, unless there are three. As for equivalence, we do not put a 9 if 
there is an equivalent 8,  do not put an 8 if 
there is an equivalent 7, and  do not put a 9 if 
there is an equivalent 7. Furthermore, we
do no put a 10, N if another sole no-value card 
N $\in \{7,8,9\}$ is left on the hand. 
Assume as a shorthand prime
variables to be of a different suit as the corresponding 
ones without primes, then we have a selection of rules of the form:
no putting of K, N' if N, 10 remain in hand;
no putting of N, N' if K, 10 remains in hand; no
putting of K, K', if N,T remains in hand;
no N, K' if K, 10 remains; no K, Q' if K, 10 remains
in hand, etc.

For suit games we forbid discarding trump cards
and apply similar constraint rules as in standard grands. We 
distinguish high- and low-trump games for putting, what we call \emph{trick-oriented} (aim at more tricks to be won) and 
\emph{eye-oriented} (aim at maximizing the number of points in a few tricks to seal the deal). In a standard trump
games, we do not allow A to be put, while in high-card trumps we do.

We have implemented a backup mechanism for the 
case 
the hard rules to choose the skat 
are over-constraint (e.g., given 11 trumps). For this case, 
some of the constraints are dropped from a second skat putting attempt.

\subsection{Soft Constraints Rules}

With soft constraints, we mean selection rules for scoring in order to construct
a ranking of the list of remaining skats.

Besides the above \emph{skat winning parameters} for the probability lookup table, we identified the following
\emph{skat scoring parameters} that 
influence good skats in trump 
games.

\begin{itemize}
    \item $f_1$: winning probability (includes  winning parameters);
    \item $f_2$: number of free suits;
    \item $f_3$: exact card value for the skat;
    \item $f_4$: number of good tens as in A, 10;
    \item $f_5$: number of bad tens as in 10, K for rear hand;
    \item $f_6$: number of 100\% standing cards (from above);
    \item $f_7$: number of standing cards with 
    retaking;
    \item $f_8$: number of standing cards without
    retaking;
    \item $f_9$: number of suits, where the declarer is leading with an ace or can retake shortly, such as 10, K.
\end{itemize}

These parameters are traded against each other in a linear  objective function
$\sum_{i=0}^9 \lambda_i \cdot f_i(h)$,
so that the skats can be ranked, and the best one being
selected.
We observe that finding the parameters $\lambda_i$
in a linear function automatically, is a prototypical machine learning problem, e.g., suitable 
for a linear classifier, or a linear SVM. 
Of course, other machine learning methods such as 
artificial neural networks or k-nearest neighbor can 
also be used to combine the features in a non-linear fashion. 

There is a snag. Putting the coefficients in 
a single linear function does not suffice.
There are simply too many 
different game sub-types that require a different 
emphasis on the skat scoring parameters. 
For example, a standard grands calls for a larger influence of standing cards, where high-card grands require a larger influence of the skat value.
There are other decisions parameters, such as the
number of As and the number of expected take-backs, dictated by the position of the player at the table, and the current selection of Js.

While setting the coefficients $\lambda_i$ for the parameters $f_i$, $i \in \{1,\ldots,9\}$, and for structuring the different 
cases to which they apply, we participate in the expertise of renown skat experts. Their long-life experience 
turned out to be more 
valuable than a learning algorithm could uncover,
even when giving thousands of training examples.

\subsection{Examples}

The trickiest example scenario for a 
refined skat selection strategy is a grand,
in particular one with two jacks. 

Grands are costly games, especially if lost,
often deciding a series or even a tournament outcome. They have to be dealt with care, as (depending on the bidding value) often there exists a 
suit game as a possible backup. In a situation without
stress to catch up, therefore, the declarer will announce a grand, only if he is almost certain that he will win. 

A worst-case 
analysis~\cite{DBLP:conf/ecai/Edelkamp20} 
to find a forced win
takes too long, so that a full game may be 
decided only in matters of an hour; and for 
skat selection we would 
need several calls to this procedure. 
Hence, we stick to expert rules such as

\begin{itemize}
    \item in forehand if at most 3 tricks are lost, put the maximum
    eyes, to play the game from above;
    \item
    do not discard cards that are 100\% standing;
    \item in rear hand with two entries, it it worth to prefer a once-supported 10 in the third suit and to put the fourth suit;
    \item if the game is won by certain, start by removing all trumps from the game;
    \item if one has to take a trick with a jack, and there are only two long suits left, try catching both other jacks is mandatory (jack jumper);
    \item if there are two aces and a third long suit, standing cards have to be built. 
\end{itemize}  

We used the following parameters
(further partition by the declarer's position, number of aces, selection of jacks, and the longest suit length).

\begin{itemize}
    \item $\lambda_1$: 2 (high-card grand), 3 (2J grand), 10 (std grand);
    \item $\lambda_2$: 2 (high-card grand), 2--25 (2J grand), 60 (std grand),
    \item $\lambda_3$: 2 (high-card grand); 3--6 (2J grand), 3 (std grand)  
    \item $\lambda_4$: 2 (high-card grand); 3 -- 6 (2J grand), 4 (std grand)   
    \item $\lambda_5$: 1 (high-card grand); 1 (2J grand), 5 (std grand)     
    \item $\lambda_6$: 2 (high-card grand); 10 (2J grand), 40 (std grand)     
    \item $\lambda_7$: 4 (high-card grand); 4 (2J grand), 40 (std grand)       
    \item $\lambda_8$: 1 (high-card grand); 41 (2J grand), 1 (std grand)      
    \item $\lambda_9$: 1 (high-card grand); 1-5 (2J grand), 1 (std grand)      
\end{itemize}

As with grand games, in suits we put the 10 in a group of 10, K, L/Q as the K
can win the trick instead; in a group of 10, K, Q 
we prefer to put 
the 10 to the Q. Even standing cards without considering retaking power should be not put easily if
others ones are available, With at least six trumps, the number of suits
should be kept small, this also works for at least four trumps, if high cards
in trump are available. In weak games with 4 six trumps and at least two non-trump
card to be lost, we also prefer to reduce the number of suits on the hand. Up
to 4 trumps one should put eyes-oriented, and from 5 trumps onwards, one should put trick-oriented. For grands with 4 Js slightly different parameters were set.

As the, the coefficients for suit games are chosen differently in high-trump
suits and low-trump suits.

\begin{itemize}
    \item $\lambda_1$: 15 (high-trump suit), 2 (low-trump suit);
    \item $\lambda_2$: 75 (high-trump suit), 35-100 (low-trump suit); 
    \item $\lambda_3$: 2 (high-trump suit), 2--4 (low-trump suit);   
    \item $\lambda_4$: 2 (high-trump suit), 20--50 (low-trump suit);      
    \item $\lambda_5$: 2 (high-trump suit), 2 (low-trump suit);      
    \item $\lambda_6$: 60 (high-trump suit), 40 (low-trump suit);      
    \item $\lambda_7$: 60 (high-trump suit), 12 (low-trump suit);         
    \item $\lambda_8$: 30 (high-trump suit), 12 (low-trump suit);         
    \item $\lambda_9$: 0 (high-trump suit), 20--25 (low-trump suit);     
\end{itemize}

In knowledge elicitation there always is a rift between the domain expert and the AI expert, whether or not a rule should lead to a hard or soft constraints. 
If you talk to a skat expert, at first s/he 
will always say that a ruling knows little or no exception, but if
you analyze the games where skats are not the best possible, then s/he might see that
omitting skats from considerations becomes questionable,
and the hard cut-off was too strong or the context to when to apply the rule 
not clear. For example, it is well-known that a high card twin with A, 10 in 
one suit wins both tricks with about 67\%, so that the 10 should better be kept
in hand (it is a standing card, too).  
On the other hand, there might be better skat candidate 
candidate, as 33\% of loosing 10 eyes is often decisive. 
Hence, it is often the case that one skat selection 
has to be seen in context to another (and the remaining hand) 
to be rejected or accepted, leading to a ranking function.

\section{Related Work}
 
The game of Skat is studied in many books~\cite{Lasker2,Wergin,Grandmontagne,Kinback,Quambush,Harmel,Rainer}. 
Kupferschmid and Helmert developed the \emph{double-dummy skat solver} (DDSS), a fast open card Skat game solver~\cite{doubledummy}.
extended to cover the partial observable game using Monte-Carlo sampling~\cite{GIB}. 
Based on DDSS, \cite{diplomkupfersc} used the LMS 
machine learning algorithm 
as a component for automated bidding. LMS is based on  
a linear evaluation function and is adopted to predict 
the expected score using 6 distinguished features of the given hand, with DDSS being called to evaluate the training set.  The algorithm is based on gradient decent on the least mean square
between the outcome of the training date and the predicted value. \cite{keller} recaps the work and
extended it to 8 features, achieving 
slightly better results. 

The bidding system of 
Keller and Kupferschmid~\cite{keller,biddingskat} is based on a $k$-nearest neighbor algorithm (KNN), also based on evaluating the training set with DDSS. The bidding module (for trump games) consists of several different evaluation functions, mapping a hand to the potential points of the game (a payoff value in $\{0, ..., 120\}$), one for each kind of game to be played. With these evaluation functions, the bidding system selects the game with the best payoff.
There are other machine learning efforts to predict bidding games and hand cards in skat~\cite{DBLP:conf/ijcai/BuroLFS09,DBLP:journals/corr/abs-1903-09604,DBLP:journals/corr/abs-1905-10907,DBLP:journals/corr/abs-1905-10911,skatfurtak}. Additionally, we have seen feature extraction in the related game of \emph{Hearts}~\cite{DBLP:conf/cg/SturtevantW06}, and automated
bidding improvements in the game of \emph{Spades}~\cite{DBLP:conf/ecai/CohensiusMOS20}. 

The bidding system of~\cite{skatfurtak} evaluates hand in conjunction with the cards that have been discarded and the type of game to be played, but instead of using a linear model they estimate winning probabilities by means of logistic regression. Basing the strength of hands on winning probability or expected payoff rather than expected card points is more suitable, because the declarers' payoff in skat mostly depends on winning the game.
Considering the bidding phase of~\cite{DBLP:journals/corr/abs-1905-10911}, deep neural nets (DNN) were trained using recorded human data from a server play. Separate networks were trained for each game type except for null and null ouvert. These were combined because of their similarity and the low frequency of ouvert games in the dataset. The features and the number of bits were manually selected and one-hot encoded. 

The results show that the prediction accuracy can be improved, however, the impact of machine learning (ML) methods to skat is not undoubted. The reported play of ML skat AIs looks promising, but according to our skat experts' opinion with experiences playing them, it still remains to be shown, on how good they actually perform.

Refering to work of  Rollason,~\cite{DBLP:conf/ecai/CohensiusMOS20} 
elaborated on
an intuitive way of statistical sampling 
the belief space of hands (worlds) based on the knowledge inferred within play. The matrices $P^i$ for the belief of card location for each player $i$ show a probability $p^i_{j,k}$ for the other players $j$ on having a card $k$ in his hand. While the approach has been developed for \emph{Spades}, it also applies to Skat. 
In a different line of research,~\cite{DBLP:conf/socs/Edelkamp19}  
showed how to predict accurate winning probabilities and play the null game. Edelkamp \cite{DBLP:conf/ecai/Edelkamp20} extended this to 
trump games, and
 studied endgame play using a complete analysis of the belief-space that is compactly kept and updated in knowledge vectors. Referring to  
combinatorial game theory~\cite{Conway}, Edelkamp \cite{DBLP:conf/ki/Edelkamp20}
proposes suit factorization and mini-game search for improved middlegame play. 

\section{Infrastructure}

Our proposed skat selection routines are implemented 
on top of the AI player bots~\cite{DBLP:conf/ecai/Edelkamp20} 
The Skat AIs themselves are chosen from a pool of clients and 
designed to support both replaying human
expert games from a file, and online interactive 
play on a server. While the clients in one pool are threads,
we often use different pools to initiate three stand-alone executables.

As indicated above,
the bidding and the game selection stages both use statistical
knowledge of winning ratios in expert games, 
stored in tables and addressed via 
patterns of winning features. They both need a predictor 
$W(h)$ for a given hand $h$ with high accuracy. 
As explained above the program estimates winning probabilities $W(h,t)$ (for hand $h$ and game of type $t$), which are extracted from the winning ratios in a database of millions of high-quality expert games. More precisely, winning probabilities $W^t(h,s)$ also including choices of skats $s$. The inferred probabilities are then exploited in the first three stages of the Skat game: bidding, skat taking and game selection, and skat putting. For each bidding value and each game type selected it generates and filters all $231$ skats and takes the average of the expected payoff of skat putting, which is determined as the maximum of the the number of $66$ skats to be put. 
The winning ratios in expert games are analyzed statistically, and by the high number of possible deals, generalizations are applied.  For null games, it estimates the winning probability $W(h,c)$ in each suit $c$. 
For all other games, it considers a hash table addressed by the 
winning parameters.
Through statistical tests~\cite{Rainer} showed that these parameters are essential attributes to accurately assess the probability of winning a trump game. In particular, a \emph{grand} table with $113,066$ entries is built on top of 7 of these winning parameters and a \emph{suit} table with $246,822$ entries using 9 of them. 
For skat putting we refined the lookup value for the different cases in a linear functions together with further winning features such as the certain/expected number of tricks (sometimes respecting the retaking options), and the exact number of points put into the skat.

Trick-taking is arranged according to an ensemble of different card recommendations. We have so-called
\begin{enumerate}
    \item \emph{killer cards} that warrant a win for the declarer/opponents to meet/violate the contract of the game -- these cards are computed by counting points for the tricks that are known to be made, including cached
    value card known to be caught from the adversaries (first 1-3 tricks), or following an extensive worst-case analysis (4-10 tricks);
    \item \emph{endgame card} as the outcome voting on the results of calling an open card game solver for the worlds that are contained in the belief space of the player with a sufficiently high confidence; 
    \item \emph{hope cards} that is the only card that can secure the game for either the declarer or the opponents, i.e., all others lead to a certain loss, this card is played instantly, the challenge is finding the hope cards by optimizing eyes or tricks.  
    \item \emph{expert cards}, for all kinds of games, position of declarer and the player within the trick the outcome of if-then-else rules that considers the hand of the players, the set of already played cards, the trick number, the (known) scores accumulated so far, the knowledge on where unseen cards might be, etc. Using human tests, the set is continuously elaborated on.
\end{enumerate}
  
Most recommendations are strict, but could be assigned to
a likelihood and confidence level.
 
\section{Experiments}

The skat selection routine is embedded in the AI~\cite{DBLP:conf/ecai/Edelkamp20} and written in C++, compiled with
\verb\gcc\ version 4.9.2 (optimization level \verb\-O2\).
Each websocket player client runs on 1 core of an
Intel Xeon Gold 6140 CPU @ 2.30GHz.
The AIs act as fully independent programs, which can play either on a server or replay games. 
We use a database of human expert games to evaluate our proposal. 
We determine the average of the game value according to the extended Seeger-System, the international agreed DSKV standard for 
evaluating game play.

There is hot debate on proper card distributions, which found its way even going to court. To avoid such trouble, the server uses an approach similar to Jan Heppe in his tool, for 
which he has been given a verified certificate. 
An alternative third option for shuffling the cards
is to draw a number in $\left[0..
{32 \choose 10}{22\choose 10}{12 \choose 10}-1\right]$ uniformly
and to unrank the position according to the following
linear-time algorithm of Myrvold and Ruskey~\cite{DBLP:journals/ipl/MyrvoldR01}.
 
 \begin{scriptsize}
\begin{lstlisting} 
unrank(n, r, pi[]) 
 if (n>0) 
   t=pi[n-1]; pi[n-1]=pi[r%n]; pi[r%n]=t;
   unrank(n-1,r/n,pi);
myrvold-ruskey(r)    
 for (int i=0;i<32;i++) pi[i]=i;
 unrank(32,r,pi);
\end{lstlisting}
\end{scriptsize}      

We side-step this discussion by replaying games
that are recorded from server play to be able
to directly compare the outcome of human play with
that of the AIs. This avoids discussing the influence 
of luck in the deal. 
While Skat is a game of skill, it
very much depends on the deal. With very good cards,
even an inferior player can beat a Skat professional.
We varied the parameters for bidding, game selection and
skat putting. This approach of evaluating
a database of played expert games, will foster 
other bots to compare, as ---due to different server technologies---
a direct cross-comparison among the bots 
is out of reach. Moreover, the players of Buro have been acknowledged to often take 5m for card selection, while 
our server imposes a timeout at 5s.
 
As depicted in Fig.~\ref{fig:skatput}, we record games 
of human players challenging our AIs on the online available
skat test server, in which the AIs are playing
against advanced club players. The results are promising, but 
as the AIs are continuously 
changing, given that humans are allowed to
drop series, where they are doomed to loose, and by the lack of an 
accepted ELO rating system, however, 
server play is less appropriate come up with a 
scientific conclusion.

\subsection{3T Grands}

We first tested our skat selection policy on 3,000 grand games with 
two jacks for the
declarer. The results in Table~\ref{tab:aigrandgames}
show that after human bidding and game selection our methods (called proposal) scores more wins (in forehand/FH, in middle hand/MH and in rearhand/RH). 
The \emph{intermediate} strategy did not properly differentiate between 
playing from above or below. If (in forehand) both Js are not won, and at most 2~tricks are going to be lost, the play is from above, if both Js fall, standing cards are recommended, even in case of having an A (to avoid opponents to drop cards to free one suit). 


The result show superior 
play wrt.\ other skat selection policies as well as to human skat putting and play.

\begin{table}[t]
    \centering
\scriptsize  
\begin{tabular}{c|ccc|c}
                        & Won FH & Won MH & Won RH & Score \\ \hline
    Proposed & {\bf 1509} & {\bf 764} & 772 &  {\bf 1251.45} \\
    Intermediate & 1499 & 762 & 771 &  1242.39 \\ \hline
    Human (Skat \& Play) & 1500 & 755 & 764 &  n.d.\\ \hline
    WinProb  & 1465 & 756 & {\bf 779} & 1193.72 \\
    Stegen & 1411 & 768 & 763 & 1151.53 \\
    Kinback & 1421 & 758 & 759 & 1141.03 \\
    Random & 1130 & 636 & 642 & 690.04 \\ \hline
    \end{tabular}
    \caption{Playing 3,300 grand games with different skat selection policies, using human bidding as well as game selection, and AI/human skat choice;
    Proposed -- refinement proposed in this paper; 
    Intermediate -- Proposed lacking some advances in play; 
    Human -- the skat the human put in the original game; WinProb -- using the ranking along winning probability table;
    Stegen -- ranking with hand strength function of Walter von Stegen; Kinback -- ranking with hand strength function of Thomas Kinback; Random -- selecting two random cards;
    the score is scaled down to 36 games and computed in the extended-Seeger scoring system (the according score for the human games has not been computed). } 
    \label{tab:aigrandgames}
\end{table}

\subsection{50T Expert Games of all Kinds}

The results on replaying 50,000 human expert games selected from good players on
our \url{32karten.de} server are presented in 
Table~\ref{tab:proposalresults}. The time for the analysis varies in between less than $44$h for games with Human bidding, game selection and skat taking, and less than $46$h for games with AI bidding and skat taking, which results in an average time of less than $3.5$s per game.
For three cases of bidding and skat putting we partition the outcome with respect to $i)$ Human game play, $ii)$ an open card solver (called Glassbox), and $iii)$ AI self play with three identical trick-taking bots (same ones in all experiments). The advances are highlighted in bold: 
we see that with the refinements and an game with our proposal,
AI bidding, game selection and skat putting
the proposal is able to achieve the highest score. 
As a sign of skill needed, a random choice 
of the two skat cards
did not even achieve half of the expected
return of the refined approaches. 

\begin{table}[t]
    \centering \scriptsize
    \begin{tabular}{c|ccc|ccc|c}
 & \multicolumn{3}{c|}{Human (Skat \& Play)}   &  \multicolumn{3}{c|}{AI (Skat \& Play)} & AI (All) \\ 
Game & Win  & Loss  & \%Won & Win  & Loss  & \%Won & \%Won \\ \hline
Suit &
26121 &
6372 &
80.4 &
{\bf 27119} &
5372 &
{\bf 83.5} &
82.3 \\ \hline
Grand &
13157 &
985&
93.0&
{\bf 13296} &
846&
{\bf 94.0} &
{\bf 95.4} \\ \hline
NO &
{\bf 1375} &
144 &
{\bf 90.5} &
{\bf 1375} &
144 &
{\bf 90.5} &
{\bf 94.9}  \\ \hline
Null  &
1187 &
659 &
64.3 &
{\bf 1327} &
519 &
{\bf 71.9} &
68.6 \\ \hline
    \end{tabular}
    \caption{Comparing human and AI playing performance in the different types of the Skat game, AI/human bidding and game selection with AI play; vs all stages AI game; NO is null
    ouvert, all sets include hand games and higher contracts. }
    \label{tab:gametypes}
\end{table}

\begin{table}[t]
    \centering
\scriptsize    \begin{tabular}{c|ccc|cc}
        & Human & Glassbox & Proposal & Won Bid & Won Bid \\ 
                & Wins & Wins & Wins &  Human  & AI\\ \hline
         Proposal  &  0 & 0 & 0 & 3886 & 2532 \\  
           &  0 & 0 & 1 & 2889 & 2445 \\
           &  0 & 1 & 0 & 166 & 186   \\
           &  0 & 1 & 1 & 1220  & {\bf 2752}  \\ 
           &  1 & 0 & 0 & 2190  & 3088  \\
           &  1 & 0 & 1 & 5637 &  5825   \\
           &  1 & 1 & 0 & 641 & 714 \\
           &  1 & 1 & 1 & 33372 & 31373   \\ \hline  
          Score &  &  &  & 951.30 &  {\bf 996.34}   \\ \hline   
         Human  &  0 & 0 & 0 &  4067 & --  \\
         (Skat)  &  0 & 0 & 1 &  {\bf 2920} & --   \\
           &  0 & 1 & 0 &  170 & --      \\
           &  0 & 1 & 1 & 1003 &  --     \\ 
           &  1 & 0 & 0 &  2034 & --    \\
           &  1 & 0 & 1 & 5500 & --      \\
           &  1 & 1 & 0 &  669  &  --   \\
           &  1 & 1 & 1 &  {\bf 33637} &  --   \\ \hline  
         Score &  &  &  & 951.53 & --      \\  \hline 
         WinProb  &  0 & 0 & 0 & 4207  &   2703   \\
           &  0 & 0 & 1 & 2695 & 2445 \\
           &  0 & 1 & 0 & 156 & 172 \\
           &  0 & 1 & 1 & 1102 & 2595  \\ 
           &  1 & 0 & 0 & 2790 & 3690 \\
           &  1 & 0 & 1 & 6259 & {\bf 6445}  \\
           &  1 & 1 & 0 & 601 & 693 \\
           &  1 & 1 & 1 & 32190 & 30171\\ \hline  
         Score &  &  &  &  917.28 &  963.56   \\  \hline 
         Stegen  &  0 & 0 & 0 & 4517 & 3067 \\  
           &  0 & 0 & 1 & 2700   &   2590  \\
           &  0 & 1 & 0 & 122    &  129     \\
           &  0 & 1 & 1 & 821    &  2129    \\ 
           &  1 & 0 & 0 & 3720   &  4746   \\
           &  1 & 0 & 1 & 8191   & 8520      \\
           &  1 & 1 & 0 & 610    & 643       \\
           &  1 & 1 & 1 &  29319    & 27090   \\ \hline  
          Score &  &  &  & 872.91 & 915.17      \\ \hline 
         Kinback  &  0 & 0 & 0 &  4570 & 3133    \\  
           &  0 & 0 & 1 &  2708  &  2620  \\
           &  0 & 1 & 0 & 123    &  127   \\
           &  0 & 1 & 1 & 759    &   2035 \\ 
           &  1 & 0 & 0 & 4202   &  5152 \\
           &  1 & 0 & 1 & 8814   & 9009 \\
           &  1 & 1 & 0 & 623    &  635 \\
           &  1 & 1 & 1 &  28202 &  26293   \\ \hline 
          Total Score &  &  &  &  852.99 &  897.52    \\ \hline  
         Random  &  0 & 0 & 0 & 6387 & -- \\  
           &  0 & 0 & 1 &  1535 & --  \\
           &  0 & 1 & 0 &  40  & -- \\
           &  0 & 1 & 1 &  198  & --   \\ 
           &  1 & 0 & 0 &  14399  & --  \\
           &  1 & 0 & 1 &  13083  & -- \\
           &  1 & 1 & 0 &   697   & -- \\
           &  1 & 1 & 1 &  13661 & --   \\ \hline  
          Total Score &  &  &  &   432.08 &
 \\ \hline 
    \end{tabular}
    \caption{Playing 50,000 mixed games with different skat selection strategies: proposed -- refinement proposed in this paper; human -- the skat the human put in the original game; WinProb -- using the ranking along winning probability table;
    Stegen/Kinback -- ranking with hand strength function of Walter von Stegen/Thomas Kinback
    (there is no
    human play for AI bidding, game and skat selection,
    so that Random and Human produced no valid results for this setting.)
    } 
    \label{tab:proposalresults}
\end{table}

With a grain of salt, we notice that for human bidding,
game selection, and skat cards being put, the AIs perform slightly 
better ($951.53$) if compared to the result of the 
same engines on human bidding,
game selection and refined skat according
to our proposal ($951.30$). 
While the Seeger score was a bit lower, 
the number of games won ($43118$, $86.23\%$) was 
larger than with the human skat ($43060$, $86.12\%$). We looked a bit
closer into the results and saw that in Grand games they are almost equivalent; we see options to reduce the remaining potential 
for further improvement in suit games.
Looking at the last column, however, where 
in an game, where we took AI bidding and game selection (in contrast
to human bidding and human game selection) as the basis, the win rates 
and scoring go up, and the latter reaches 
a superior performance
of $996.34$ points in the extended Seeger scoring system (scaled to 
a series of 36 games). 
With AI bidding, $2.1\%$ foldings (no player bids $18$) were noticed.

Looking closer at the data, it is quite obvious that AI skat putting fits 
better to the AI bidding than to the human one, 
given that both are using the same 
trick-taking algorithms. 
One reason is that the notion
of a \emph{good skat}, which is a skat with a scoring of at most $20\%$ off,
in AI play, fits better to a skat selected with the bot.
As Table~\ref{tab:gametypes} highlights, the AIs with human bidding and game selection 
has better winning ratios than humans in their play, for all types of games, except null ouvert (where there is a tie), even when taking the human skat. With respect to
the winning probability we also see 
a significant positive effect for the refinement proposal 
on the playing strength measured in the extended Seeger score.  
Note that ---depending on the largest bid--- 
both null and null-ouvert are often chosen in time of
trouble, when finding a bad skat, 
to reduce the expected Seeger score to be lost.



By changing the evaluation formula in Equation~\ref{eq:eval} (substituting the second $50$ with $90$), we could increase the mean extended Seeger-score over the bar of $1000$ at the price of more folded games (we achieved $1012$ Seeger points on average per $36$ games), but the number of foldings rose from $2.2\%$ to $4.6\%$). Conversely, we
could reduce the number of foldings to less than $2\%$ for a slightly smaller Seeger score of $993.2$ (instead of $996.34$). This allows to
adapt to different bidding levels of bidding aggressiveness of the players, which, by using winning probabilities, we can derive from the database (a simple form of opponent modeling). 

In $5363$ games the AIs played grand, while the humans suit with $90.5\%$ wins, against only $2643$ games, where
the AIs play suit and the human play grand. This shows that the AIs are better in going in for a higher game than
the humans, leading to a score that was $29\%$ higher.
From $8000$ games the humans folded, the AIs were able to bid
$5112$, of which $4211$ were won with an average score of
$820.31$.
  
\subsection{83T Games of all Kinds}  

We performed another comparison on $83848$ expert games we got 
from \url{www.euroskat.com}. 
The results are shown in Table~\ref{tab:proposal80results}.
The AI won $(4184+4770)-(5132+1048)=2774$ more games than 
the humans in their own games. Also, we see that AI bidding and 
skat selection makes up to about 45 extended Seeger points on average, 
in this case leading to an average score of more than 1000.
In the bidding process, the AIs decided to fold $83848-81900=1948$ 
games ($2.32\%$).

\begin{table}[t]
    \centering
\scriptsize    \begin{tabular}{c|ccc|cc}
        & Human & Glassbox & Proposal & Won Bid \\ 
                & Wins & Wins & Wins  &\\ \hline
         Proposed  
         &  0 & 0 & 0 & 4021  \\  
         Skat    &  0 & 0 & 1 & 4184  \\
           &  0 & 1 & 0 & 302    \\
           &  0 & 1 & 1 & 4770   \\ 
           &  1 & 0 & 0 & 5132    \\
           &  1 & 0 & 1 & 9567    \\
           &  1 & 1 & 0 & 1048  \\
           &  1 & 1 & 1 & 52876    \\ \hline  
          Score &  &  &  & {\bf 1005.07}   \\ \hline   
         Human  &  0 & 0 & 0 & 6574   \\
         Skat  &  0 & 0 & 1 & 5180       \\
           &  0 & 1 & 0 & 293        \\
           &  0 & 1 & 1 & 1666     \\ 
           &  1 & 0 & 0 & 3377     \\
           &  1 & 0 & 1 & 9272      \\
           &  1 & 1 & 0 & 1103  \\
           &  1 & 1 & 1 & 56382    \\ \hline  
         Score &  &  &  & 959.77       \\  \hline  
    \end{tabular}
    \caption{Playing $83848$ games of all kinds with different skat selection strategies: (tob) all refinement including AI bidding; (bottom) 
    the skat the human put in the original game.  
   } 
    \label{tab:proposal80results}
\end{table}

\section{Conclusion}

Skat selection is a critical decision 
for the game of Skat, and fully deserves to coin its name.
With hard and soft constraint rules, we provided an advanced system for selecting appropriate skats. The results show that our refined mechanism is superior to many alternative skat putting strategies for trump games on top of the prediction of winning probabilities.

We replayed human games, 
to avoid counterbalancing the influence of chance 
in the deal, which for the game of Skat is widely experienced to be large.
Although the winning probabilities were accurate, they are not sufficient to determine good skats 
to beat human experts. 

The coefficients $\lambda_i$ for the additionally proposed winning parameters for skat selection are manually set. 
of games using the winning ratio or point scoring result, given that manual tuning is cumbersome, even if it pays 
off. We defer this automation to future research.
The combinatorial problem to face is that different hands call for different weights, and that in total there are several winning parameters already. 

\paragraph*{Acknowledgement}
We thank Rainer G\"o{\ss}l for his invaluable 
contribution of expert-level Skat play and 
Stefan Meinel for his annotated play and mathematical insights.

\bibliographystyle{IEEEtran}
\bibliography{select.bib}

\end{document}